\documentclass[preprint,3p,times,authoryear]{elsarticle}

\usepackage{amssymb}
\usepackage{amsthm}
\usepackage{amsmath}
\usepackage{amssymb}
\usepackage{scalefnt}
\usepackage{hyperref}
\usepackage{scalefnt}
\usepackage{multirow}
\usepackage{setspace}

\journal{Elsevier}

\begin{document}

\begin{frontmatter}



\title{Paragraph-based complex networks: application to document classification and authenticity verification}

\author[label1]{Henrique F. de Arruda}
\author[label1]{Vanessa Q. Marinho}
\author[label2]{Luciano da F. Costa}
\author[label1,label3]{Diego R. Amancio}
\ead{diegoraphael@gmail.com}
\address[label1]{Institute of Mathematics and Computer Science, University of S\~ao Paulo, S\~ao Carlos, SP, Brazil.}
\address[label2]{S\~ao Carlos Institute of Physics, University of S\~ao Paulo, S\~ao Carlos, SP, Brazil.}
\address[label3]{School of Informatics, Computing and Engineering, Indiana University, Bloomington, Indiana 47408, USA.}

\begin{abstract}
With the increasing number of texts made available on the Internet, many applications have relied on text mining tools to tackle a diversity of problems. A relevant model to represent texts is the so-called word adjacency (co-occurrence) representation, which is known to capture mainly syntactical features of texts.In this study, we introduce a novel network representation that considers the semantic similarity between paragraphs. Two main properties of paragraph networks are considered: (i) their ability to incorporate characteristics that can discriminate real from artificial, shuffled manuscripts and (ii) their ability to capture syntactical and semantic textual features. Our results revealed that real texts are organized into communities, which turned out to be an important feature for discriminating them from artificial texts. Interestingly, we have also found that, differently from traditional co-occurrence networks, the adopted representation is able to capture semantic features. Additionally, the proposed framework was employed to analyze the Voynich manuscript, which was found to be compatible with texts written in natural languages. Taken together, our findings suggest that the proposed methodology can be combined with traditional network models to improve text classification tasks.
\end{abstract}


\end{frontmatter}

\section{Introduction}

Due to the ever-increasing number of available online texts, many machine learning techniques have been developed to treat this kind of information~\cite{manning2008introduction,stamatatos2009survey,pak2010twitter,yaverouglu2014revealing,symeonidis2018comparative,sicilia2018twitter,xiong2018deep}. Among many statistical methods, network-based approaches have also been proposed to address several natural language processing problems, including writing style analysis~\cite{10.1371/journal.pone.0118394}, authorship attribution~\cite{mehri2012complex,amancio2011comparing} and sentiment analysis~\cite{zhao2014sentiment}. Several graph-based approaches hinge on the topological information of the obtained networks to perform some type of classification~\cite{i2001small,liu2013language,amancio2015complex,erkan2004lexrank,angelova2006graph,jin2007graph,yu2017hybrid}.

A well-known representation of texts as complex networks 
is the co-occurrence model~\cite{i2001small,liu2013language,amancio2015complex,wachs2016analyzing}. This model represents words as nodes, and edges are established for every pair of adjacent words. Recently, this representation was found to capture mainly syntax features~\cite{amancio2013probing,masucci2006network}, which has been confirmed by numerous works using co-occurrence networks to study language styles~\cite{segarra2015authorship,masucci2006network,de2016using,mehri2012complex,amancio2015comparing,cong2014approaching}. In order to grasp features that go beyond syntax, other models have been proposed. In~\cite{de2016topic}, the authors still consider words as nodes, but the connections are created considering a larger window, rather than only consecutive words. Upon applying community detection methods, this approach was successfully employed to detect topics. Regarding the mesoscopic scale, a network based on similarity of large chunks was proposed in~\cite{ferraz2017representation}. This methodology was found to be useful to understand and visualize the unfolding of stories~\cite{marinho2017Calligraphy}. In the summarization context, another approach that also took into consideration larger chunks of texts is the network of connected paragraphs~\cite{salton1997automatic}.

In this work, we propose a novel paragraph-based network, which takes into consideration textual similarity by employing \emph{tf-idf} (term frequency-inverse document frequency) weighting~\cite{Manning1999Language} together with cosine similarity. Differently from previous approaches~\cite{salton1997automatic}, the paragraph-based networks considered here are analyzed in terms of their topological and dynamical properties. The properties of the adopted network representation were probed by considering two different criteria. To test the informativeness of the networks, we investigated whether paragraph-based networks are able to discriminate real from shuffled texts. In the second test, we analyzed if the networks are able to capture syntax and, mostly importantly, semantic textual information. Our results showed that the modularity played an important role in distinguishing real and shuffled texts, since the presence of communities turned out to be a characteristic inherent of real texts. We also found that particular measurements are able to capture semantic features of texts, a feature that has not been observed in most co-occurrence networks modeling texts~\cite{amancio2013probing}. 

In addition to the analysis aimed at better understanding the statistical properties of paragraph-based networks, we probed the statistical properties of an unknown text -- the Voynich manuscript -- using the framework proposed here. Differently from other approaches, we did not assume that pages are organized in any specific order. This is an important feature because a recent study revealed that the traditionally assumed pages ordering might be unreliable~\cite{reddy2011we}. Interestingly, our results indicate that the Voynich manuscript is compatible with natural languages and incompatible with shuffled texts.  These conclusions were mostly corroborated by observing the community structure arising from the manuscript.

The remainder of this paper is organized as follow. In section~\ref{sec:mat}, we present the employed datasets, the proposed methodology, and the used complex network measurements. Section~\ref{sec:results} presents the an analysis of the paragraph-network properties by comparing real documents with two versions of shuffled texts. Furthermore, in the same section, we present a case study where we analyze the Voynich manuscript. Finally, in Section~\ref{sec:conc}, we conclude the study with perspectives for further works. 

\section{Materials and methods} \label{sec:met}

\label{sec:mat}
This section describes the employed datasets, the approach devised to create paragraph-based networks and the measurements extracted from the text networks.

\subsection{Dataset}

We employed two datasets. The first one, henceforth referred to as the Holy Bible dataset, was used to represent the variation of syntax \emph{across different languages} when the text/content is the same.  It comprises three books from the New Testament of the Holy Bible: Matthew, Mark and Luke. $16$ different languages were considered: Arabic, Basque, English, Esperanto, German, Greek, Hebrew, Hungarian, Korean, Latin, Maori, Portuguese, Russian, Swahili, Vietnamese, and Xhosa. The three books were concatenated into a single document so as to obtain a larger text, as our method is more reliable when larger pieces of texts are used to construct the network. This same procedure has been applied in similar studies~\cite{amancio2013probing}. For all considered languages, the paragraphs comprise the same verses. In total, $658$ paragraphs were manually identified.

The second dataset, henceforth referred to as Books dataset, comprises $53$ books in different languages, namely English, French, German, Italian and Portuguese. This dataset was used to analyze how the network structure varies \emph{across different documents} in the {same language}.
The list of books is presented in Appendix \ref{apendice}.

\subsection{Paragraph-based networks}

In this work, texts are modeled as complex networks. A network (or graph) can be defined as a set $V = \{v_1, v_2, ..., v_n\}$ of nodes and a set $E = \{e_1, e_2, ..., e_m\}$ of edges. In a unweighted network, the element $a_{ij}$ of the adjacency matrix $\mathbf{A}$ is equal to 1 if node $i$ is connected to node $j$; otherwise, $a_{ij} = 0$. In weighted networks, the element $a_{ij}$ corresponds to the weight of the link between nodes $i$ and $j$.

\begin{figure}[!htpb]
  \centering
     \includegraphics[width=0.65\textwidth]{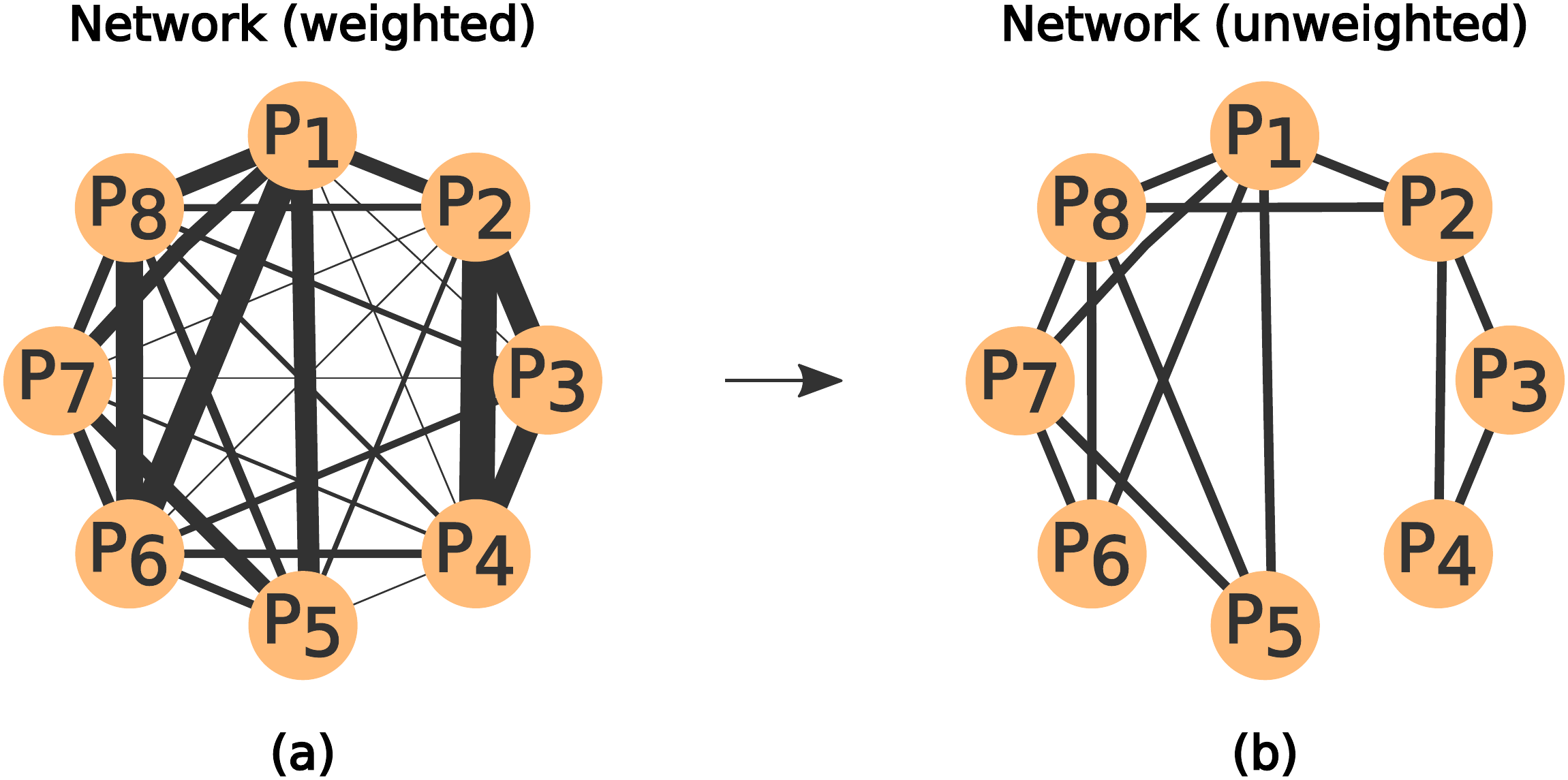}
   \caption{Example of thresholding paragraph-based networks. In the weighted version (a), all nodes (paragraphs) are connected among themselves and the weight of each edge is given by the textual (semantical) similarity between the nodes. The unweighted version (b) is obtained by removing all edges with weights below a given threshold \emph{T}.}
  \label{fig:paragraph}
\end{figure}

The main objective of the adopted network model is to represent how short contexts (i.e. paragraphs) semantically relate to each other in a textual document. 
To create a paragraph-based network, the raw text is divided into chunks of paragraphs. Each paragraph is considered as a network node, as illustrated in Figure~\ref{fig:paragraph}(a)). In order to establish links between paragraphs, each node is considered as a document $d$ in the set $D$ of documents. The \emph{tf-idf} (term frequency-inverse document frequency) weighting map~\cite{Manning1999Language} is then computed to quantify the relevance of each word $w \in d$:
\begin{equation}
  \text{tf-idf}(w,d,D) = \frac{f_{w,d}}{n} \times \log\Bigg{(} \frac{|D|}{d_w}\Bigg{)},
\end{equation}
where $f_{w,d}$ is the frequency of $w \in d$, $n$ is the total number of words in $D$ and $d_w$ is the number of documents (paragraphs) in which $w$ appears. For each paragraph, a vector containing the {tf-idf} weights for the words is created and then edge weights are computed by using the cosine similarity for all pairs of paragraphs (nodes). Note that this methodology creates a fully connected, weighted graph, as illustrated in Figure~\ref{fig:paragraph}(a). Because many of the complex networks measurements are defined only for unweighted networks, we removed the weakest edges using a threshold $T$. For the considered networks, we chose a threshold for each network in order to keep all networks with the same size and density $E=5\%$. This is an important step in the pre-processing phase because several network measurements are known to be very sensitive to both size and density~\cite{costa2007characterization,10.1371/journal.pone.0118394}. In preliminary experiments, we found that a perturbation in $T$ does not alter the conclusions reported here. The effect of thresholding the weighted network is illustrated in Figure~\ref{fig:paragraph}(b).

The proposed methodology is similar to the mesoscopic networks approach~\cite{ferraz2017representation} in terms of the network edge weights. Actually, paragraph networks can be understood as a specific case of mesoscopic networks in which each chunk of text is a single paragraph with no forced overlap between adjacent chunks. An advantage of the present study is that here we can analyze texts in which the order of the pages and paragraphs are unknown. Note that other techniques of word representation, such as word embeddings, were not considered here because the proposed method was developed to be applied even in texts whose language is unknown.

\subsection{Network variations}

In order to compare real and shuffled texts, three types of networks were considered. The paragraph-based network -- denoted as \emph{real texts} (RT) -- is obtained from the pre-processed texts of the considered datasets, as described in Section \ref{sec:met}. The other networks are obtained from shuffled versions of the original text. The versions are created by shuffling words (\emph{SW}) or sentences (\emph{SS}). 
The versions obtained from an extract of the book \emph{The Adventures of Sherlock Holmes}, by Arthur Conan Doyle are:

\begin{enumerate}
\item \emph{Real Text (RT) version}: "Quite so," he answered, lighting a cigarette, and throwing himself down into an armchair. "You see, but you do not observe. The distinction is clear. 

\item \emph{Shuffled words (SW) version}: "Quite a into do distinction armchair. but lighting and answered, The observe. himself down you so," not throwing he see, cigarette, is clear. "You an  

\item \emph{Shuffled Sentences (SS) version}: "You see, but you do not observe. "Quite so," he answered, lighting a cigarette, and throwing himself down into an armchair. The distinction is clear.
\end{enumerate}

\subsection{Network characterization}
The following network measurements were used to characterize the paragraph-based networks:

\begin{enumerate}

\item {\textbf{Degree ($k$):}} This measurement quantifies the number of immediate neighbors of a node $i$~\cite{costa2007characterization} and it is obtained as $k_{i} = \sum_{j} A_{ij}$.

\item \textbf{{Betweenness ($B$):}}
This measurement quantifies the relevance of a node (or edge) in terms of the number of shortest paths including that node (or edge)~\cite{Boccaletti2006}. The betweenness centrality of a given node $i$ is calculated as 
\begin{equation}
B_{i} = \sum_{s,t} \frac{g^{i}_{s,t}}{g_{s,t}} ,
\end{equation}
where $g^{i}_{s,t}$ is the number of shortest paths connecting nodes $s$ and $t$ that include node $i$, and $g_{s,t}$ is the number of shortest paths connecting $s$ and $t$, for all pairs $s$ and~$t$. In text networks, this measurement has been applied to identify if a concept/node is semantically related to one or more topological communities~\cite{amancio2011comparing}.

\item \textbf{{Clustering coefficient (cc):}} The clustering coefficient represents the probability of two neighbors of a given node being connected with each other~\cite{costa2007characterization}. Locally, the clustering coefficient is calculated as $cc_i = 2e_i/(k_i^2 - k_i)$. In text analysis, the clustering coefficient has also been used to identify if a concept appears in generic or specific contexts. Differently from the betweenness, only local information is considered.

\item \textbf{{Neighborhood ($N$):}} this measurement quantifies the amount of nodes in the $h$-th concentric level around node $i$~\cite{Newman2010}. In this study, we used $h = 3$. 
              
\item {\textbf{Eccentricity (Ecc):}} the eccentricity of a node $i$ is a centrality index equal to the maximum length of all the shortest paths from $i$ to the other nodes in the network~\cite{Harary69a}.
              
\item \textbf{{Eigenvector centrality (EC):}} the eigenvector centrality assigns a value to a given node $i$ proportional to the sum of the eigenvector centrality values of the nodes connected to $i$. By doing so, the centrality value of a node increases when it is connected to nodes with high eigenvector centrality~\cite{Newman2010}.
              
\item \textbf{{Closeness centrality ($C$):}} this measurement is given by the inverse of the average distance from a node to the other nodes in the network~\cite{Newman2010}. It is obtained as
$C_i = l_i^{-1} = n/\sum_{j}{d_{ij}}$, where ${l_i}$ is the average distance from node $i$ to all the other nodes, and $d_{ij}$ is the length of a geodesic path connecting nodes $i$ and $j$. 
              
\item \textbf{{Accessibility ($\alpha^{(h)}$):}} This measurement quantifies the number of accessible nodes at the $h$-th concentric level centered at node $i$~\cite{Travencolo2008} (we used $h = \{2,3\} $). This analysis accounts for the accessibility of a node taking into account the probability $p_{i,j}^{(h)}$ of a random walker to reach a given node $j$ departing from $i$, in $h$ steps. The equation that describes this measurement is based on the Shannon entropy, as follows
\begin{equation}
\alpha^{(h)}_i = \exp{\left(-\sum{p_{i,j}^{(h)} \log p_{i,j}^{(h)}} \right)}.
\end{equation}
In language networks, the accessibility (and its variations) has been used as an important feature to identify the relevance of words in the context of structural/stylistic analysis~\cite{10.1371/journal.pone.0118394,amancio2015complex}.

\item \textbf{{Generalized Accessibility ($\alpha^{(\infty)}$):}} The generalized accessibility does not depend on the parameter $h$. In contrast with the previous measurement, generalized accessibility uses a modified random walk, called accessibility random walk, which assigns higher weights to the shortest paths and penalizes the longest ones~\cite{generalized_acessibility}. Mathematically, the measurement is defined as 
\begin{equation}
\alpha^{(\infty)}_i = \exp{ \left(-\sum{P_{i,j} \log P_{i,j}} \right)},
\end{equation}
where $P$ is computed as the probability transition of all the pairs of nodes $i$ and $j$. More details are available in~\cite{generalized_acessibility}.

\item \textbf{{Symmetry ($S^{(h)}$):}} As another variation of accessibility, this measurement quantifies the symmetry of the topology around a given node $i$, by considering its neighborhood ($h$)~\cite{Silva}. $S^{(h)}$ is defined in a two-fold manner: (i) the \emph{backbone} ($Sb^{(h)}$), in which the connections between nodes in the same hierarchical level ($h$) are removed and (ii) \emph{merged} ($Sm^{(h)}$), where the nodes that are connected and belong to the same hierarchical level ($h$) are merged into a single node. The measurement is computed as %
\begin{equation}
S^{(h)}_i = \frac{\exp{\left(-\sum{p_{i,j}^{(h)} \log p_{i,j}^{(h)}} \right)}}{|H_h(i)| + \sum_{r=0}^{h-1}{\eta_r}},
\end{equation}
where $H_h(i)$ is the set of all nodes  in the $h-th$ hierarchic level of node $i$, $|H_h(i)|$ is the number of nodes in $H_h(i)$, and by considering a given hierarchic level $r$, $\eta_r$ is the number of nodes without edges connecting to the next hierarchical level. In this study, we employed $h = \{2,3,4\}$. In text networks, the symmetry has been useful to identify the authorship of texts~\cite{2015concentric}.
             
\item \textbf{{Modularity ($Q$):}} proposed by~\citet{NewGir04}, the modularity measures the quality of a given network partitioning in terms of its communities. It can be obtained as:
\begin{equation}
Q = \frac{1}{2m}\sum_{i=1}^{n}\sum_{j=1}^{n}\bigg[ a_{ij} - \frac{k_i k_j}{2m}\bigg]\delta (c_i,c_j),
\end{equation}
where $m$ is the number of edges, $n$ is the number of nodes, $\delta (c_i,c_j) = 1$ if the nodes $i$ and $j$ are from the same class (community) and $\delta (c_i,c_j) = 0$, otherwise. This measurement ranges from $-\infty \leq Q < 1$. For $Q > 0$, the number of edges inside the communities is greater than the expected in a equivalent random network. In other words, a positive value of modularity is an indicative that the network is organized in communities.
\end{enumerate}

Apart from the modularity, all of the aforementioned measurements are locally defined, i.e. each node has a specific value. To summarize the values obtained for a measurement $X$ across all nodes of the network, we took the average ($\langle X \rangle$) and the standard deviation ($\sigma(X)$). Note that this approach has already been adopted in similar works~\cite{de2016using,marinho2017Calligraphy}.

An important issue arising from  the characterization and classification of networks concerns the comparison of networks with different sizes. Since several network measurements may depend on the total number of nodes, we decide to construct the networks so as the total number of nodes (paragraphs) is constant. 

\subsection{Informativeness analysis}

In the adopted network representation, we define as informative the measurements whose values obtained from real books and the respective shuffled versions are significantly different. Measures complying with this condition are therefore able of discriminating between real and random manuscripts. Note that an informative measurement is useful to verify if an unknown manuscript is compatible with a known textual structure (e.g. the structure observed in documents written in natural languages).

Two criteria were used to test the informativeness of the networks:
\begin{enumerate}

    \item \emph{Criterion A:} this criterion is aimed at verifying if the values obtained from the set of all shuffled texts of the dataset can be discriminated from the values obtained for all real texts. Let $N_\textrm{RT}$ and $N_\textrm{S}$ be the total number of books in the RT dataset and the number of shuffled versions generated for each book in RT, respectively. Here, we perform a comparison of $N_\textrm{RT}$ values in RT with $N_\textrm{RT} \cdot N_\textrm{S}$ values obtained from shuffled texts.    

    \item \emph{Criterion B:} it consists in comparing the value obtained for the real (RT) text with the values obtained in the corresponding shuffled versions of the same text.
    For a given measurement $X$, the distance between a real text and the respective shuffled versions is obtained by computing the \emph{z-score} (i.e. the standard score):
    \begin{equation} \label{eq:zes}
        z(X) = \frac{x - \langle X^{(R)} \rangle}{\sigma (X^{(R)})},
    \end{equation}
    where $x$ is the value obtained in the real text, $X^{(R)}$ is the set of values obtained from the $N_\textrm{S}$ shuffled versions (SW or SS); and $\langle \ldots \rangle$ and $\sigma$ represent the mean and standard deviation of the distribution, respectively.
    
\end{enumerate}

In our tests, for each real text, we created $N_\textrm{S} = 30$ samples for both SW and SS versions.

\subsection{Dependency with language and semantics}

An important property to be verified in a text network is the ability of the extracted measurements to capture syntactical and/or semantical features of the represented texts~\cite{amancio2013probing}. In order to study the dependency of the measurements on syntax and semantics, the measurements are extracted in two classes of datasets. For a given measurement, $X_{t=\pi,l}$ represents the set of values obtained for $X$ in a dataset comprising the same book $t=\pi$ in different languages ($l$). In a similar fashion, $X_{t,l=\lambda}$ represents the set of values obtained for $X$ in a dataset comprising different texts ($t$) written in the same language $l=\lambda$. If a given network measurement depends more on the language (i.e. the syntax) than on the approached subject (i.e. the semantics), one expects that variability of the distribution of $X_{t=\pi,l}$ will be larger than the variability of $X_{t,l=\lambda}$. Conversely, if $X$ is more dependent on  semantics, one expects that the variability of $X_{t,l=\lambda}$ will be larger than the variability of $X_{t=\pi,l}$~\cite{amancio2013probing}.
Here, the  variability of the distributions is computed by using the coefficient of variation (CV) of the distribution, i.e. 
\begin{equation}    \label{eq.cv}
    \textrm{CV}(X) = \frac{\sigma (X)}{\langle X \rangle}.
\end{equation}

\section{Results and discussion}

\label{sec:results}
In this section, we analyze the properties of the metrics extracted from  the proposed network representation. Here we focus on two main properties: informativeness and the ability of the metrics to capture  syntactical and/or semantic textual features. The applicability of the adopted representation is then illustrated in the analysis of an unknown text: the Voynich manuscript.

\subsection{Informativeness}

In this study, we used distinct ways to quantify informativeness~\cite{amancio2013probing}. In the first approach, we consider a measurement $X$ as informative if the value obtained for $X$ in a real (RT) text differs from the values of $X$ obtained in \emph{any other} shuffled (SW and SS) text of the considered dataset (see \emph{Condition A} described in the methodology).
The results obtained for this type of analysis are shown in Table~\ref{tab:all}. To facilitate the comparison of measurements taking values in distinct intervals, a normalization was applied. For each measurement, and for each of the datasets (Holy Bible and Books), the results are standardized considering all three types of texts (RT, SS and SW). As such, the average value of each normalized measurement in Table~\ref{tab:all} is zero (i.e. $\textrm{RT}+\textrm{SS}+\textrm{SW}=0$) and the standard deviation is one. 

\begin{table*}[!tpb]
\centering
\begin{tabular}{|l|c|c|c|c|c|c|}
\hline
\bf  &  \multicolumn{3}{c|}{\bf Holy Bible dataset} & \multicolumn{3}{c|}{\bf Books dataset (all)} \\ \hline
\bf $X$ & \bf RT & \bf SS & \bf SW & \bf RT & \bf SS & \bf SW \\ \hline
$\langle k \rangle$  		& $ -0.12 \pm 1.14$ & $  +0.13 \pm 0.87$ & $ -0.01 \pm 0.96$ & $ -0.18 \pm 1.02$ & $ +0.05 \pm 0.98$ & $ +0.13 \pm 0.98$ \\ \hline
$\sigma (k)$ 		 		& $ -0.59 \pm 0.52$ & $  +0.56 \pm 1.15$ & $ +0.02 \pm 0.86$ & $ -0.78 \pm 0.65$ & $ +0.47 \pm 0.92$ & $ +0.31 \pm 0.90$ \\ \hline
$\langle B \rangle$  		& $ +0.48 \pm 0.90$ & $-0.23 \pm 1.02$ & $ -0.24 \pm 0.90$ & $ -0.05 \pm 1.07$ & $ +0.05 \pm 1.03$ & $ +0.00 \pm 0.89$ \\ \hline
$\sigma (B)$ 		 		& $ -0.22 \pm 0.89$ & $  +0.23 \pm 1.11$ & $ -0.01 \pm 0.93$ & $ -0.66 \pm       0.68$ & $       +0.52 \pm       0.94$ & $       +0.13 \pm       0.97$ \\ \hline
$\langle cc \rangle$ 		& $  +0.74 \pm 0.60$ & $-0.58 \pm 1.09$ & $ -0.16 \pm 0.74$ & $  +0.31 \pm       0.90$ & $     -0.17 \pm       1.06$ & $     -0.15 \pm       0.96$ \\ \hline
$\sigma (cc)$ 		 		& $  +0.52 \pm 0.45$ & $-0.32 \pm 1.37$ & $ -0.20 \pm 0.70$ & $-0.28 \pm       0.64$ & $       +0.27 \pm       1.21$ & $       +0.01 \pm       0.98$ \\ \hline
$\langle N \rangle$  		& $ -0.14 \pm 0.74$ & $-0.09 \pm 1.33$ & $   +0.23 \pm 0.76$ & $  +0.31 \pm       0.83$ & $     -0.13 \pm       1.14$ & $     -0.18 \pm       0.92$ \\ \hline
$\sigma (N)$ 		 		& $  +0.39 \pm 0.76$ & $-0.18 \pm 1.28$ & $ -0.21 \pm 0.75$ & $-0.28 \pm       0.85$ & $       +0.08 \pm       1.13$ & $       +0.20 \pm       0.94$ \\ \hline
$\langle Ecc \rangle$		& $  +0.64 \pm 1.04$ & $-0.42 \pm 0.91$ & $ -0.22 \pm 0.67$ & $  +0.01 \pm       1.08$ & $     -0.09 \pm       1.03$ & $       +0.07 \pm       0.87$ \\ \hline
$\sigma (Ecc)$ 		 		& $  +0.11 \pm 1.02$ & $  +0.01 \pm 1.09$ & $ -0.13 \pm 0.86$ & $-0.38 \pm       1.04$ & $       +0.18 \pm       0.93$ & $       +0.20 \pm       0.92$ \\ \hline
$\langle EC \rangle$ 		& $  +0.73 \pm 1.22$ & $-0.54 \pm 0.49$ & $ -0.19 \pm 0.64$ & $-0.18 \pm       1.13$ & $       +0.06 \pm       0.92$ & $       +0.12 \pm       0.91$ \\ \hline
$\sigma (EC)$ 		 		& $  +0.33 \pm 1.09$ & $-0.13 \pm 0.96$ & $ -0.20 \pm 0.85$ & $-0.53 \pm       0.82$ & $       +0.32 \pm       1.01$ & $       +0.21 \pm       0.93$ \\ \hline
$\langle C \rangle$  		& $ -0.64 \pm 0.79$ & $  +0.32 \pm 1.02$ & $   +0.33 \pm 0.84$ & $-0.02 \pm       1.08$ & $       +0.00 \pm       1.03$ & $       +0.02 \pm       0.89$ \\ \hline
$\sigma (C)$ 		 		& $ -0.42 \pm 0.55$ & $  +0.43 \pm 1.28$ & $ -0.01 \pm 0.83$ & $-0.87 \pm       0.59$ & $       +0.44 \pm       0.89$ & $       +0.43 \pm       0.84$ \\ \hline
$\langle Sb^{(2)} \rangle$  & $ -0.09 \pm 0.67$ & $-0.09 \pm 1.26$ & $   +0.18 \pm 0.96$ & $  +0.48 \pm       0.87$ & $     -0.40 \pm       0.97$ & $     -0.09 \pm       0.96$ \\ \hline
$\sigma (Sb^{(2)})$ 		& $ +0.13  \pm 0.53$ & $  +0.06 \pm 1.39$ & $ -0.19 \pm 0.84$ & $-0.45 \pm       0.69$ & $       +0.29 \pm       1.13$ & $       +0.16 \pm       0.96$ \\ \hline
$\langle Sm^{(2)} \rangle$  & $  +0.12  \pm 0.37$ & $-0.33 \pm 1.52$ & $   +0.21 \pm 0.63$ & $  +0.55 \pm       0.83$ & $     -0.34 \pm       1.06$ & $     -0.20 \pm       0.85$ \\ \hline
$\sigma (Sm^{(2)})$ 		& $ +0.18  \pm 0.60$ & $-0.02 \pm 1.20$ & $ -0.16 \pm 1.08$ & $  +0.37 \pm       0.62$ & $     -0.43 \pm       1.24$ & $       +0.06 \pm       0.87$ \\ \hline
$\langle Sb^{(3)} \rangle$  & $ +0.13  \pm 0.65$ & $-0.20 \pm 1.29$ & $   +0.06 \pm 0.93$ & $  +0.41 \pm       0.90$ & $     -0.26 \pm       1.02$ & $     -0.14 \pm       0.94$ \\ \hline
$\sigma (Sb^{(3)})$ 		& $-1.09  \pm 0.68$ & $  +0.61 \pm 0.57$ & $   +0.48 \pm 0.65$ & $-0.52 \pm       0.96$ & $       +0.21 \pm       1.00$ & $       +0.31 \pm       0.82$ \\ \hline
$\langle Sm^{(3)} \rangle$  & $-0.48  \pm 0.72$ & $  +0.51 \pm 1.16$ & $ -0.03 \pm 0.81$ & $-0.38 \pm       1.01$ & $       +0.15 \pm       1.01$ & $       +0.23 \pm       0.85$ \\ \hline
$\sigma (Sm^{(3)})$ 		& $-0.66  \pm 0.96$ & $  +0.53 \pm 0.89$ & $   +0.13 \pm 0.74$ & $-0.46 \pm       1.05$ & $       +0.15 \pm       0.94$ & $       +0.30 \pm       0.83$ \\ \hline
$\langle Sb^{(4)} \rangle$  & $+0.30  \pm 0.61$ & $-0.02 \pm 1.39$ & $ -0.28 \pm 0.72$ & $-0.28 \pm       0.90$ & $       +0.11 \pm       1.10$ & $       +0.17 \pm       0.93$ \\ \hline
$\sigma (Sb^{(4)})$ 		& $+0.42  \pm 0.62$ & $-0.25 \pm 1.29$ & $ -0.18 \pm 0.83$ & $-0.30 \pm       0.91$ & $     -0.01 \pm       1.11$ & $       +0.31 \pm       0.87$ \\ \hline
$\langle Sm^{(4)} \rangle$  & $+1.00  \pm 0.71$ & $-0.69 \pm 0.60$ & $ -0.30 \pm 0.76$ & $+0.35 \pm       1.10$ & $     -0.28 \pm       0.93$ & $     -0.07 \pm       0.84$ \\ \hline
$\sigma (Sm^{(4)})$ 		& $-0.23  \pm 0.95$ & $  +0.24 \pm 1.05$ & $ -0.01 \pm 0.94$ & $-0.57 \pm       1.07$ & $       +0.29 \pm       0.83$ & $       +0.28 \pm       0.82$ \\ \hline
$\langle \alpha^{(\infty)} \rangle$ & $ -0.22 \pm 1.10$            & $ -0.08 \pm 1.02$ & $+0.30 \pm 0.78$ &  $       +0.31 \pm       0.95$ & $     -0.25 \pm       1.04$ & $     -0.06 \pm       0.93$ \\ \hline
$\sigma (\alpha^{(\infty)})$ 		& $ -0.20 \pm 1.00$ & $ -0.25 \pm 0.86$ & $  +0.45 \pm 0.98$ &  $     -0.86 \pm       0.71$ & $      +0.37 \pm       0.85$ & $      +0.49 \pm       0.82$ \\ \hline
$\langle \alpha^{(2)} \rangle$ 		& $ -0.50 \pm 0.52$ & $   +0.12 \pm 1.28$ & $  +0.38 \pm 0.82$ &  $       +0.19 \pm       0.96$ & $     -0.17 \pm       1.05$ & $     -0.02 \pm       0.96$ \\ \hline
$\sigma (\alpha^{(2)})$ 			& $ -0.04 \pm 0.81$ & $ -0.15 \pm 1.16$ & $  +0.19 \pm 0.97$ &  $     -0.79 \pm       1.05$ & $      +0.20 \pm       0.65$ & $      +0.60 \pm       0.68$ \\ \hline
$\langle \alpha^{(3)} \rangle$ 		& $  +0.02 \pm 0.51$ & $ -0.24 \pm 1.38$ & $  +0.22 \pm 0.86$ &  $       +0.48 \pm       0.78$ & $     -0.33 \pm       1.05$ & $     -0.15 \pm       0.96$ \\ \hline
$\sigma (\alpha^{(3)})$ 			& $  +1.03 \pm 0.56$ & $ -0.78 \pm 0.68$ & $-0.26 \pm 0.69$ &  $       +0.77 \pm       1.01$ & $     -0.52 \pm       0.78$ & $     -0.25 \pm       0.68$ \\ \hline
$Q$ 								& $  +1.32 \pm 0.33$ & $ -0.91 \pm 0.22$ & $-0.41 \pm 0.31$ &  $       +1.24 \pm       0.65$ & $     -0.64 \pm       0.38$ & $     -0.60 \pm       0.37$ \\ \hline

\end{tabular}
\caption{Measurements obtained for the different network types (RT, SS, and SW) by considering the Holy Bible dataset, the English part of the book dataset, and the entire book dataset (see Appendix~\ref{apendice}). Note that all the presented data is standardized to be possible to compare different measurements.}
\label{tab:all}
\end{table*}

Considering the Holy Bible dataset, the modularity ($Q$) was the measurement that best discriminated real from shuffled texts. The modularity in real networks differs $2.24$ and $1.73$ from the SW and SS versions, respectively. This result suggests that the community structure is much more apparent in real networks, which might be a consequence of the bursty topical textual structure present in real texts~\cite{de2016topic}. 
In addition to the modularity, other measurements were also found to be informative. When comparing RT and SW, the largest differences of values were found for the accessibility ($\sigma (\alpha^{(3)})$), symmetry ($\sigma (Sb^{(3)})$ and $\langle Sm^{(4)} \rangle$) and the clustering coefficient ($\langle cc \rangle$). The best discrimination between  RT and SS was found for the symmetry ($\sigma (Sb^{(3)})$ and $\langle Sm^{(4)} \rangle$), accessibility ($\sigma (\alpha^{(3)})$) and closeness ($\langle C \rangle$). Interestingly, 
several of the measurements were able to distinguish between real and shuffled texts, regardless of the considered shuffling process.

Considering the Books dataset, the modularity also turned out to be the measurement that best discriminated real from shuffled texts. Once again, real texts oftentimes displayed a clearer community structure. This means that the informativeness achieved by the modularity is a characteristic that seems to depend neither on syntax nor semantics.
Apart from $Q$, the following measurements were also found to discriminate real from both shuffled networks: clustering coefficient ($\sigma (C)$),  degree ($\sigma (k)$) and accessibility ($\alpha^{(\infty)}$, and $\sigma (\alpha^{(2)})$ and $\sigma (\alpha^{(3)})$).

As a complementary test, for each measurement, we used the \emph{z-score} (see equation \ref{eq:zes}) to compare a real text and its corresponding shuffled versions (informativeness test based on \emph{Condition B}). Note that this is a less strict informativeness test because, differently from the previous case, we do not compare a real text with shuffled versions from \emph{all texts} of the dataset. Here, we rather compare a real text and the shuffled versions generated only from the \emph{same book}. 
In Table~\ref{tab:zscore}, we show the percentage of documents in which we observed a significant difference between real and shuffled texts -- according to the z-score defined in equation \ref{eq:zes}. 

\begin{table*}[]
\centering
\begin{tabular}{|l|c|c|c|c|c|}
\hline
\multicolumn{2}{|c|}{\multirow{2}{*}{\bf Measurements}} & \multicolumn{2}{c|}{\bf SW} & \multicolumn{2}{c|}{\bf SS} \\
\cline{3-6}
\multicolumn{2}{|c|}{} &  \bf Holy Bible & \bf Books &  \bf Holy Bible & \bf Books \\
\hline 
\multirow{2}{*}{Degree}& $\langle k \rangle$ & $31.25\%$ & $30.30\%$ &  $18.75\%$ & $36.36\%$ \\ 
& $\sigma (k)$ & $56.25\%$ & $78.79\%$ & $31.25\%$ & $72.73\%$ \\ \hline
\multirow{2}{*}{Betweenness} & $\langle B \rangle$ & $37.50\%$ & $48.48\%$ &  $50.00\%$ & $57.58\%$ \\
& $\sigma (B)$ & $6.25\%$ & $66.67\%$ & $6.25\%$ & $60.61\%$ \\ \hline
\multirow{2}{*}{Clustering} & $\langle cc \rangle$ & $50.00\%$ & $27.27\%$ &  $50.00\%$ & $39.39\%$ \\ 
& $\sigma (cc)$ & $12.50\%$ & $30.30\%$ & $62.50\%$ & $30.30\%$ \\ \hline
\multirow{2}{*}{Neighborhood}& $\langle N \rangle$ & $6.25\%$ & $42.42\%$ & $6.25\%$ & $63.64\%$ \\
& $\sigma (N)$ & $25.00\%$ & $39.39\%$ & $37.50\%$ & $60.61\%$ \\ \hline
\multirow{2}{*}{Eccentricity} & $\langle Ecc \rangle$ & $37.50\%$ & $30.30\%$ & $43.75\%$ & $45.45\%$ \\
& $\sigma (Ecc)$ & $37.50\%$ & $51.52\%$ & $56.25\%$ & $54.55\%$ \\ \hline
\multirow{2}{*}{Eigenvector} & $\langle EC \rangle$ & $62.50\%$ & $57.58\%$ & $62.50\%$ & $57.58\%$ \\
& $\sigma (EC)$ & $31.25\%$ & $78.79\%$ & $37.50\%$ & $66.67\%$ \\ \hline
\multirow{2}{*}{Closeness}  & $\langle C \rangle$ & $43.75\%$ & $39.39\%$ & $62.50\%$ & $51.52\%$ \\
& $\sigma (C)$ & $12.50\%$ & $90.91\%$ & $25.00\%$ & $93.94\%$ \\ \hline
\multirow{12}{*}{Symmetry} & $\langle Sb^{(2)} \rangle$ & $6.25\%$ & $54.55\%$ &  $25.00\%$ & $51.52\%$ \\
& $\sigma (Sb^{(2)})$ & $      6.25\%$ & $     42.42\%$ & $     12.50\%$ & $     42.42\%$ \\
& $\langle Sm^{(2)} \rangle$ & $0.00\%$ & $57.58\%$ & $6.25\%$ & $63.64\%$ \\
& $\sigma (Sm^{(2)})$ & $6.25\%$ & $30.30\%$ & $12.50\%$ & $42.42\%$ \\
& $\langle Sb^{(3)} \rangle$ & $18.75\%$ & $45.45\%$ & $18.75\%$ & $48.48\%$ \\
& $\sigma (Sb^{(3)})$ & $93.75\%$ & $36.36\%$ & $93.75\%$ & $48.48\%$ \\
& $\langle Sm^{(3)} \rangle$ & $56.25\%$ & $57.58\%$ & $25.00\%$ & $66.67\%$ \\
& $\sigma (Sm^{(3)})$ & $75.00\%$ & $57.58\%$ & $68.75\%$ & $57.58\%$ \\
& $\langle Sb^{(4)} \rangle$ & $0.00\%$ & $54.55\%$ & $37.50\%$ & $66.67\%$ \\
& $\sigma (Sb^{(4)})$ & $18.75\%$ & $63.64\%$ & $37.50\%$ & $66.67\%$ \\
& $\langle Sm^{(4)} \rangle$ & $93.75\%$ & $54.55\%$ & $87.50\%$ & $54.55\%$ \\
& $\sigma (Sm^{(4)})$ & $31.25\%$ & $54.55\%$ & $31.25\%$ & $66.67\%$ \\ \hline
\multirow{6}{*}{Accessibility} & $\langle \alpha^{(\infty)} \rangle$ & $0.00\%$ & $54.55\%$ & $37.50\%$ & $54.55\%$ \\
& $\sigma (\alpha^{(\infty)})$ & $12.50\%$ & $93.94\%$ & $18.75\%$ & $93.94\%$ \\
& $\langle \alpha^{(2)} \rangle$ & $12.50\%$ & $42.42\%$ & $50.00\%$ & $42.42\%$ \\
& $\sigma (\alpha^{(2)})$ & $18.75\%$ & $51.52\%$ &  $12.50\%$ & $81.82\%$ \\
& $\langle \alpha^{(3)} \rangle$ & $0.00\%$ & $54.55\%$ & $25.00\%$ & $51.52\%$ \\
& $\sigma (\alpha^{(3)})$ & $100.00\%$ & $66.67\%$ & $87.50\%$ & $63.64\%$ \\ \hline
Modularity & $Q$ & $100.00\%$ & $100.00\%$ & $100.00\%$ & $100.00\%$ \\ 
\hline
\end{tabular}
\caption{Percentage of documents in each dataset where the difference between a real text (English part) and the corresponding shuffled version was found to be significant. Apart from the modularity, the informativeness seems to depend on the type of dataset used. }
\label{tab:zscore}
\end{table*}

As found in the first test, $Q$ is the most critical measurement for both of the considered datasets, reaching 100\% of informativeness. Other measurements had similar results for both datasets, eg., $\sigma(k)$, $\langle Sm^{(3)} \rangle$, and $\langle cc \rangle$, which we found to be informative for approximately 50\% of the samples. However, for many other measurements, the level of informativeness varied according to the dataset. For example, $\sigma (Sb^{(3)})$ and $\sigma (\alpha^{(3)})$ were found to be more informative in the Holy Bible dataset. Conversely, $\sigma (C)$ and $\langle Sm^{(2)} \rangle$ seemed to be more informative in the Books dataset.

All in all, the results obtained here suggest that, apart from the modularity, it is important to analyze the characteristics of the dataset to decide if network measurements extracted from paragraph networks can be classified as informative -- even if a less strict definition of informativeness is taken into account. Interestingly, the results obtained here confirm that paragraph networks are less informative than other types of text networks~\cite{10.1371/journal.pone.0118394}. In the case of word adjacency networks, most of the measurements were found to be informative, independently of the characteristics of the considered datasets.

\subsection{Dependency on syntax and semantics}

In this section, we evaluate the dependency of the measurements by considering their variability in two distinct scenarios: in datasets where (i) the semantics (text) is constant and the language (syntax) is varied; and (ii) the language is constant and the semantics varies. To represent (i), we used the Holy Bible dataset. The dataset employed in the second scenario was created by selecting only the Books in English from the Books dataset. We decided to use the English language because, in the considered dataset, a larger number of books written in this language is available. 

In the first analysis, we identified the measurements that were able to capture syntax/language subtleties. The measurements that were found to display significant variability in this scenario (i.e. in the Holy Bible dataset) were: accessibility ($\langle \alpha^3 \rangle$ and $\sigma (\alpha^2)$), degree ($\langle k \rangle$), eccentricity ($\sigma (Ecc)$), symmetry ($\langle Sb^2 \rangle$, $\langle Sb^3 \rangle$, $\sigma (Sm^4)$ and $\sigma (Sb^2)$), neighborhood ($\langle N \rangle$) and betweenness ($\sigma (B)$).

We also identified the measurements that are sensitive to changes in semantics. The measurements taking the highest coefficients of variation in the English Books datasets were: eccentricity ($\langle Ecc \rangle$), closeness ($\langle C \rangle$), symmetry ($\langle Sm^4 \rangle$ and $\langle Sb^4 \rangle$), betweenness ($\langle B \rangle$), degree ($\langle k \rangle$), eigenvector centrality ($\langle EC \rangle$), accessibility ($\langle \alpha^2 \rangle$ and $\langle \alpha^\infty \rangle$) and neighborhood ($\langle N \rangle$). Note that some measurements might depend on both syntax and semantics. This is the case of $\langle N \rangle$. Interestingly, for both symmetry and accessibility, the ability to capture syntax or semantics subtleties depends on the hierarchical level being analyzed. 

\begin{figure*}[!htbp]
 \centering
 \includegraphics[width=1.\textwidth]{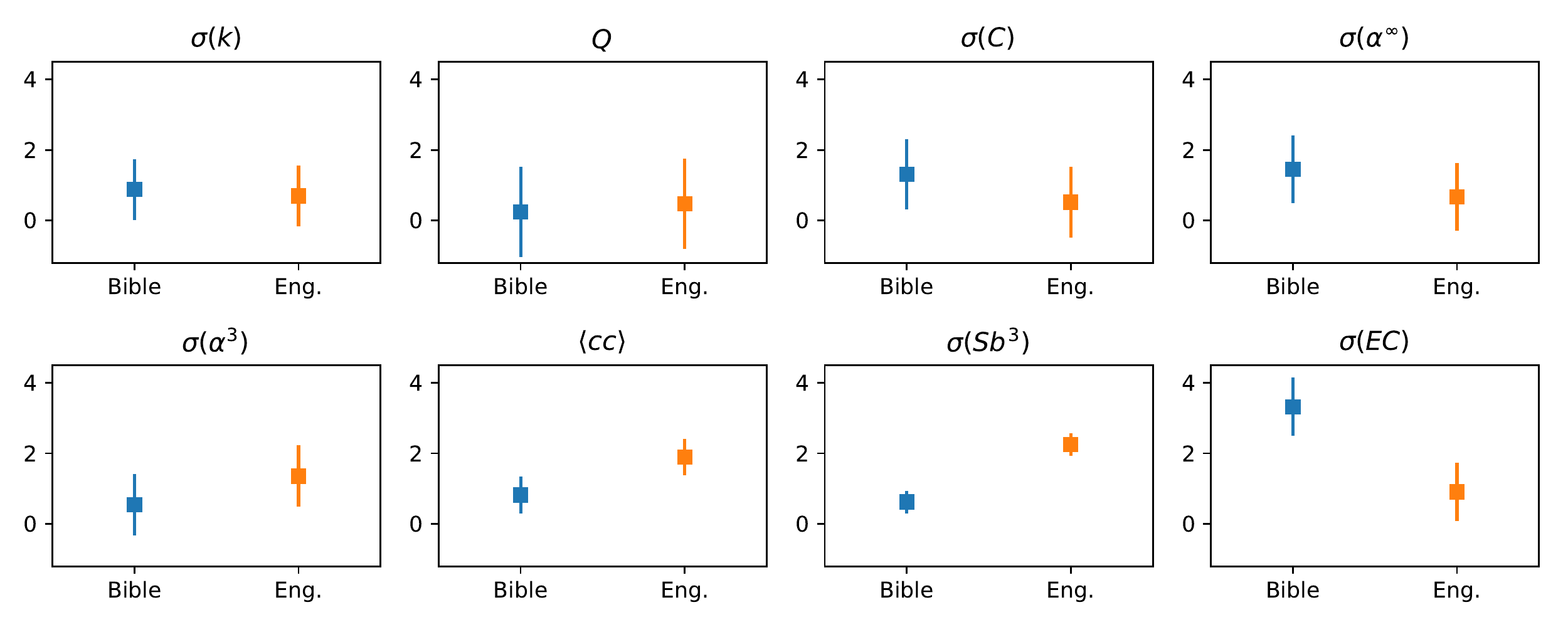}
 \caption{Comparisons of the measurements that provided lower values CV. The error bar represents the confidence interval of a mean for the interval of 95\% of confidence.}
 \label{fig:cv}
\end{figure*}

In addition to the aforementioned tests, we probed, for each measurement, which of the two phenomena is more prevalent: (i) the ability to detect changes in syntax; and (ii) the ability to detect changes in semantics. This prevalence analysis was conducted by comparing the coefficient of variation in the considered datasets, as described in the methodology. The obtained results are shown in Figure \ref{fig:cv}. The top sub-panels illustrate the results obtained for $\sigma(k)$, $Q$, $\sigma(C)$ and $\sigma(\alpha^\infty)$. In most of these cases, while the variability across languages (Bible dataset) or topics (English dataset) is high, there is no significant difference between these values. This means that, for these measurements, both syntax and semantics are captured.

A different behavior can be observed for the measurements depicted in the bottom sub-panels of Figure \ref{fig:cv}. For both $\langle cc \rangle$ and $\sigma(Sb^{(3)})$, a significant difference of coefficients of variation was found. For the first three measurements, the variability across topics turned out to be significantly higher than the variability across topics. This is an interesting finding in text networks, since measurements extracted from other texts networks (such as co-occurrence networks) are mostly dependent on syntax~\cite{amancio2013probing}. This result suggests that paragraph-based networks can be used to complement the analysis based on traditional co-occurrence networks when both syntax and semantics are relevant for the problem being addressed.

\subsection{Classification tests}

To illustrate the applicability of the paragraph-based network in classification tasks, some classification problems were tackled using the . In the first example, we considered the problem of deciding whether a manuscript has a structure compatible with a shuffled, meaningless document. In the second classification problem, we probed whether an unknown text -- the Voynich manuscript -- can be considered compatible with real texts.

\subsubsection{Discriminating real and shuffled texts}

We applied our method to distinguish real from shuffled texts in order to illustrate the capabilities of paragraph based networks to characterize texts regarding a real application. For each book presented in Appendix~\ref{apendice}, the three paragraph-based networks were created, RT, SW, and SS. After that, the network measurements described previously were extracted, the values were standardized, and those values were used as classification features. To select the features for this task, we considered the most informative measurements obtained from Table~\ref{tab:all}. More specifically, for each pair real vs. shuffled texts (i.e., RT vs. SS and RT vs. SW) we identified the top 10 measurements providing the best discrimination. Then, we selected those measurements appearing in both top 10 lists.
The measurements selected are: $Q$, $\sigma (C)$, $\sigma (\alpha^{(\infty)})$, $\sigma (k)$, $\sigma (B)$, $\sigma (\alpha^{(3)})$, $\sigma (\alpha^{(2)})$, $\langle Sm^{(2)} \rangle$, and $\sigma (EC)$.


The classification was evaluated by the live-one-out cross-validation and the SMO classifier algorithm, which is an SVM implementation available in Weka~\cite{witten2016data,hall2009weka}. The parameters were chosen according to the procedure defined in~\cite{amancio2014systematic}. When considering three classes (RT, SW, and SS), the accuracy was $76.08\%$. However, the true positive rate was 0.98 for the RT samples and 0.71 and 0.50 for SW and SS, respectively. The false negative rates ware 0.02, 0.24, and 0.14 for RT, SW, and SS, respectively. These results mean that the proposed framework can easily differentiate between real and shuffled texts. Conversely, the discrimination  between the two classes of shuffled documents represent a more challenging task.

\begin{figure}[!htbp]
 \centering
 \includegraphics[width=0.5\textwidth]{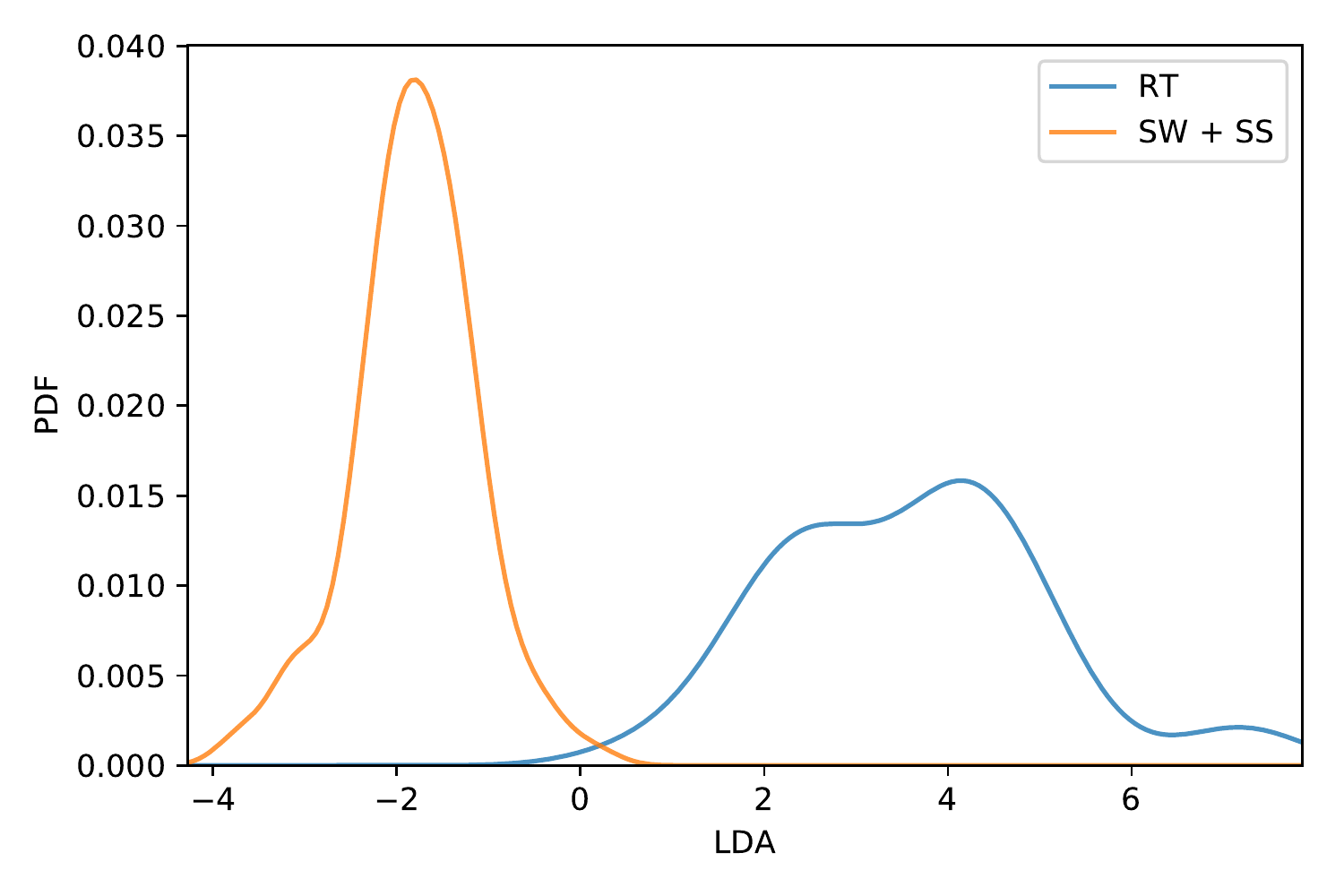}
 \caption{Probability density function (pdf) of the linear discriminant analysis projection obtained from the selected features in the classification of texts in two classes: real vs. shuffled (SW and SS) texts.}
 \label{fig:lda}
\end{figure}

A variation of the same classification problem considered both shuffled versions as being the same class. In this case, SVM reached  $98.72\%$ of accuracy. The false positive rate of the RT networks was 0, and the only two classification mistakes were made by real texts classified as shuffled texts. Figure~\ref{fig:lda} illustrates the separation between the two classes by considering the projection into a single dimension obtained via linear discriminant analysis~\cite{friedman2001elements}. Given the importance stressed by the modularity in the informativeness analysis, we also evaluated the performance when only this measurement is used for the classification. In this case, the accuracy rate reached $96.79\%$, which confirms the importance of the modularity in discriminating real and shuffled texts.

\subsubsection{Case Example: Voynich manuscript}

The Voynich manuscript is known to be a mysterious text, and many of its aspects have been studied for several years~\cite{reddy2011we}. Some studies have relied on textual analysis~\cite{reddy2011we}, while others have used complex networks tools to study its properties~\cite{amancio2013probing,montemurro2013keywords}. 
In order to handle the manuscript -- originally written in an unknown alphabet -- it is necessary to translate its characters into a known set of symbols.
Here we used the European Voynich Alphabet (EVA)~\cite{voynich}, which provides  the original characters  manually translated into European characters. To provide a better quality translation, for each line of the text, different translations are available. Here we considered the voting of the most recurring character for all different translations of the same line. Additionally, because our approach relies on text paragraphs, we detected  paragraphs by visually inspecting  the original manuscript.
When comparing the Voynich manuscript with shuffled texts, we disregarded
the SS versions because there is no trivial way to detect sentences in the Voynich manuscript.

\begin{figure*}[!htbp]
 \centering
 \includegraphics[width=0.99\linewidth]{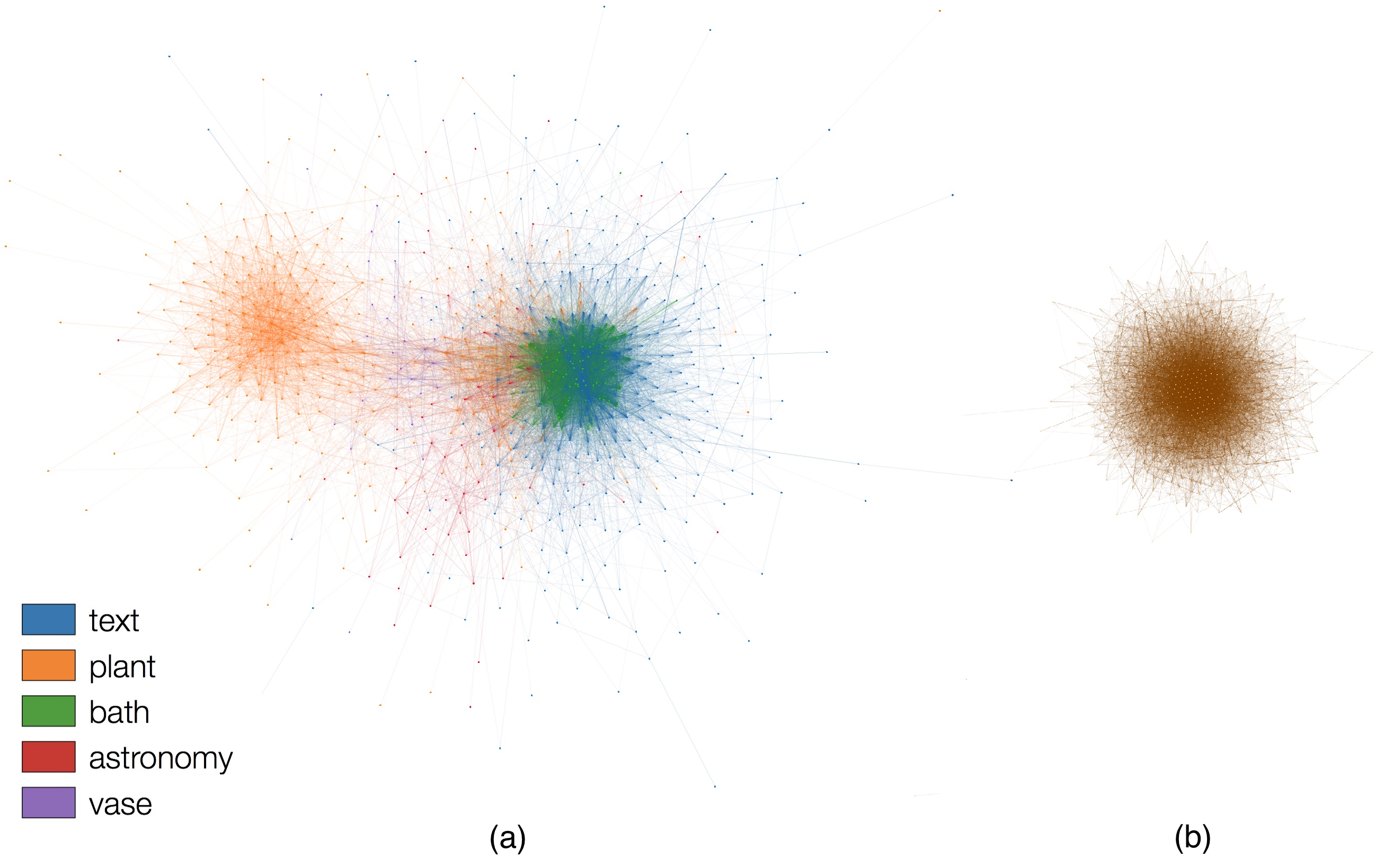}
 \caption{Network visualizations of the two version of paragraph based networks of Voynich manuscript. In (a) the paragraphs were labelled by considering the figures in the corresponding pages. The topics considered were: (i) \emph{text}, when no images are available; (ii) \emph{plants}; (iii) \emph{bath}, with figures of women and bath-like shape; (iv) \emph{astronomy}, with spatial-like figures; and (v) \emph{vase}. The visualization was provided by the software implemented by~\cite{silva2016using}.}
 \label{fig:visual}
\end{figure*}

First, we analyzed if the Voynich manuscript, when characterized with the metrics extracted from paragraph-based networks, is compatible with real texts and not compatible with gibberish, shuffled texts. This is a long-standing question about the manuscript, since several scholars have questioned the existence of a meaningful textual structure in this mysterious text~\cite{belfield2007six}. An illustration comparing the structure of the Voynich manuscript and a shuffled network is shown in Figure~\ref{fig:visual}. It is clear from the visualizations that the Voynich manuscript presents an well-defined community structure, with two dominant groups. The communities seems to capture the topical organization of the manuscript in some degree: the extract about plants seems to be separated in a specific community. 
The equivalent shuffled network, shown in Figure~\ref{fig:visual}(b) reveals no apparent community structure. Since the modularity was found to be informative in the previous analysis, the organization in communities in Figure~\ref{fig:visual}(a) suggests, at the paragraph level, that  the Voynich manuscript is not compatible with shuffled texts. Interestingly, this same conclusion has been reported when different types of networks are used to represent the manuscript~\cite{amancio2013probing,montemurro2013keywords,reddy2011we}.

\begin{figure}[!htpb]
  \centering
    \includegraphics[width=0.5\textwidth]{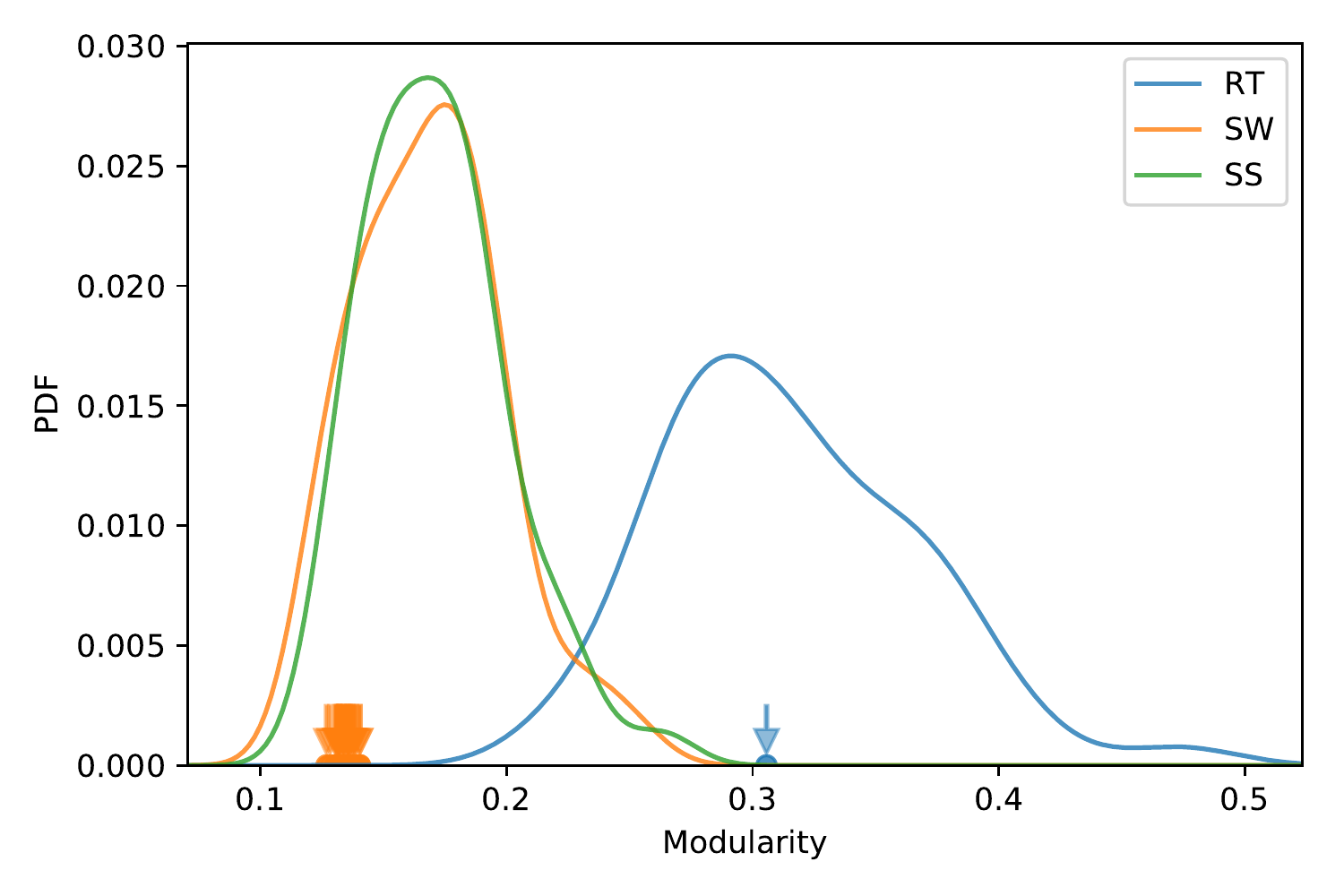}
   \caption{Probability density function (pdf) of the modularity measurement for  three types of networks (RT, SW, and SS). The modularity of the two versions of the Voynich manuscript RT and SW are represented by the blue and orange arrows, respectively.}
  \label{fig:modularity}
\end{figure}

As a complementary analysis, taking into account the community structure of the networks, we analyzed the modularity $Q$, which is much higher for the set of real texts (RT) when compared to the set of shuffled texts (SS and SW), as shown in Figure~\ref{fig:modularity}. 
The modularity obtained for the Voynich (represented with a blue arrow in the figure) is not compatible with any of the two distributions obtained for shuffled texts.  On the other hand, the modularity of the manuscript is compatible with the modularity extracted from real texts. 

In order to analyze the Voynich manuscript, we employed the same classifier as in the previous section. As a result, the document was classified as real text. A set of 30 SW networks of Voynich were also classified, and the accuracy of 100\% was found. This perfect classification can also be seen in Figure \ref{fig:modularity}, which shows that the generated SW networks of Voynich (orange arrows) are mostly compatible with the distributions obtained for shuffled texts.

\section{conclusions}
\label{sec:conc}

In the current study, we probed the properties of a paragraph-based networked representation of texts. Two main properties were considered: the ability of the networks to distinguish real from shuffled texts (informativeness test) and the ability to capture syntactic and/or semantic text features. Interestingly, we found that the most informative measurement is the modularity, since artificial, shuffled texts are not organized in well-defined communities. Our results also revealed that several measurements are able to capture semantic features. This is an important feature, since the well-known word adjacency (co-occurrence) networks are only able to capture syntax features. Our findings suggest that both co-occurrence and paragraph-based networks can be used in a complementary way when both syntax and semantics important for a natural language processing task.

The adopted network representation was used to analyze the statistical nature of the Voynich manuscript. Previous studies hinging on word networks showed that the Voynich syntax is coherent with natural languages~\cite{amancio2013probing,montemurro2013keywords}. Recently, an extensive analysis using several natural languages argued that Hebrew is the most probable language of the manuscript~\cite{hauer2016decoding}. Here, we proposed a different analysis, by focusing on the organization in paragraphs. Our analysis revealed that the Voynich manuscript is compatible with natural languages at the paragraph level. This finding was confirmed  by analyzing the organization of the text into well-defined communities: similarly to several natural languages, the Voynich also displays a clear community structure organization. Furthermore, we applied our classification approach and the Voynich manuscript was classified as real text. As a complement, the accuracy of $100\%$ was found when we classified 30 samples of the shuffled version of the Voynich manuscript.

As a future work, many other natural language problem can be addressed by considering the proposed network. For instance,
several problems currently being addressed by word adjacency models could be benefited from the paragraph-based network approach. Examples of applications include machine translation quality, analysis of plagiarism and authorship attribution~\cite{amancio2015comparing}. Additionally, other unknown documents can also be examined in terms of their organization in paragraphs~\cite{belfield2007six}.

\section*{Acknowledgments}
Henrique F. de Arruda acknowledges Capes-Brazil for sponsorship.  Vanessa Q. Marinho thanks FAPESP (grant no. 2015/05676-8) for financial support. Luciano da F. Costa thanks CNPq (grant no. 307333/2013-2) and NAP-PRP-USP for sponsorship. Diego R. Amancio acknowledges FAPESP (grant no. 16/19069-9 and 17/13464-6) for financial support. This work has been supported also by FAPESP grants 11/50761-2 and 2015/22308-2. The authors acknowledge Filipi Nascimento Silva for fruitful conversations.

\renewcommand{\thetable}{A\arabic{table}}    

\setcounter{table}{0}  

\newpage

\appendix

\section{Dataset} \label{apendice}

The list of books used to analyze how the network structure varies across different documents in the \emph{same language} is shown below. Five different languages were considered: English, French, German, Italian and Portuguese. The list of books is organized by language. The author of each book is listed between parentheses after each title. The books were obtained from the Project Gutenberg~\footnote{http://www.gutenberg.org}.

\begin{enumerate}

    \item {\bf English:} The Adventures of Sherlock Holmes (Arthur Conan Doyle), The Tragedy of the Korosko (Arthur Conan Doyle), The Valley of Fear (Arthur Conan Doyle), Uncle Bernac - A Memory of the Empire (Arthur Conan Doyle), Dracula's Guest (Bram Stoker), The Lair of the White Worm (Bram Stoker), The Jewel Of Seven Stars (Bram Stoker), The Man (Bram Stoker), The Mystery of the sea (Bram Stoker), A Tale of Two Cities (Charles Dickens), Barnaby Rudge: A Tale of the Riots of Eighty (Charles Dickens), American Notes (Charles Dickens), Great Expectations (Charles Dickens), Hard Times (Charles Dickens), The Works of Edgar Allan Poe -- Volumes 2 and 4 (Edgar Allan Poe), Beasts and Super-Beasts (Hector H. Munro), The Chronicles of Clovis (Hector H. Munro), The Toys of Peace (Hector H. Munro), The Girl on the Boat (P. G. Wodehouse), My Man Jeeves (P. G. Wodehouse), Something New (P. G. Wodehouse), The Adventures of Sally (P. G. Wodehouse), The Clicking of Cuthbert (P. G. Wodehouse), A Pair of Blue Eyes (Thomas Hardy), Far from the Madding Crowd (Thomas Hardy), Jude the Obscure (Thomas Hardy), The Mayor of Casterbridge (Thomas Hardy), The Hand of Ethelberta (Thomas Hardy), Barry Lyndon (William M. Thackeray), The History of Pendennis (William M. Thackeray), The Virginians (William M. Thackeray) and Vanity Fair (William M. Thackeray).
    
    \item {\bf French:} Le fils du Soleil (Gustave Aimard), Face au Drapeau (Jules Verne), Pierre de Villergl\'e (Louis Am\'ed\'ee Achard), Les Idoles d'argile (Louis Reybaud) and Han d'Islande (Victor Hugo).
    
    \item {\bf German:} Die Wahlverwandtschaften (Goethe), Der Moloch (Jakob Wassermann), K\"onigliche Hoheit (Thomas Mann) and Lichtenstein (Wilhelm Hauff).
    
    \item {\bf Italian:} Il Peccato di Loreta (Alberto Boccardi), La Montanara (Anton Giulio Barrili),  Alla Finestra (Enrico Castelnuovo), Sciogli la treccia, Maria Maddalena (Guido da Verona) and La Pergamena Distrutta  (Virginia Mulazzi).
    
    \item {\bf Portuguese:} Amor de Perdição  (Camilo Castelo Branco), A Cidade e as Serras (E\c{c}a de Queir\'os), Os Bravos do Mindello (Faustino da Fonseca), Transviado (Jaime de Magalh\~aes Lima) and Uma Fam\'ilia Inglesa (J\'ulio Dinis).

\end{enumerate}

\bibliographystyle{elsarticle-harv}
\bibliography{references.bib}

\end{document}